%% file: main.tex
\documentclass{article}

\usepackage[preprint]{neurips_2024}

\usepackage[utf8]{inputenc} % allow utf-8 input
\usepackage[T1]{fontenc}    % use 8-bit T1 fonts
\usepackage{hyperref}       % hyperlinks
\usepackage{url}            % simple URL typesetting
\usepackage{booktabs}       % professional-quality tables
\usepackage{amsfonts}       % blackboard math symbols
\usepackage{nicefrac}       % compact symbols for 1/2, etc.
\usepackage{microtype}      % microtypography
\usepackage{xcolor}         % colors
\usepackage{amsmath,amssymb}
\usepackage{wrapfig}
\usepackage{graphicx}
\usepackage[toc,page]{appendix}
\usepackage{float}

\title{Keras Sig: Efficient Path Signature Computation on GPU in Keras 3}

\author{%
  Rémi Genet\\
  DRM\\
  Université Paris Dauphine\\
  Paris, France \\
  \texttt{remi.genet@dauphine.psl.eu} \\
  \And
  Hugo Inzirillo \\
    Finance-Insurance\\
  CREST-IP Paris \\
  Palaiseau, France \\
  \texttt{hugo.inzirillo@ensae.fr} \\
}

\begin{document}

\maketitle
\begin{abstract}
In this paper we introduce \textit{Keras Sig} a high-performance pythonic library designed to compute path signature for deep learning applications. Entirely built in Keras 3, \textit{Keras Sig} leverages the seamless integration with the mostly used deep learning backends such as PyTorch, JAX and TensorFlow. Inspired by \cite{kidger2021},we proposed a novel approach reshaping signature calculations to leverage GPU parallelism. This adjustment allows us to reduce the training time by 55\% and 5 to 10-fold improvements in direct signature computation compared to existing methods, while maintaining similar CPU performance. Relying on high-level tensor operations instead of low-level C++ code, \textit{Keras Sig} significantly reduces the versioning and compatibility issues commonly encountered in deep learning libraries, while delivering superior or comparable performance across various hardware configurations. We demonstrate through extensive benchmarking that our approach scales efficiently with the length of input sequences and maintains competitive performance across various signature parameters, though bounded by memory constraints for very large signature dimensions. 
\end{abstract}

\input{1_Introduction}

\input{2_Innovation}

\input{3_Performance_Analysis}

\input{4_Conclusion}
\bibliographystyle{plainnat}
\bibliography{bib}
\end{document}

%% file: 1_introduction.tex
\section{Introduction}

The signature transform, a central object in rough path theory \cite{lyons1998, lyons2014}, has emerged as a powerful tool for feature extraction in machine learning tasks involving sequential data. By capturing nonlinear interactions and temporal dependencies in data streams, the signature transform has been successfully applied to a wide range of applications, including time series analysis \cite{kidger2020}, handwriting recognition \cite{yang2016}, and medical diagnostics \cite{morrill2019}. Despite its theoretical appeal, the practical adoption of the signature transform in machine learning has been hindered by computational challenges, particularly in scaling to large datasets and integrating with modern deep learning frameworks. Existing libraries for computing signatures, such as \textit{Signatory} \cite{kidger2021}, have made significant strides in optimizing these computations. However, these implementations often rely on low-level C++ code and are tightly coupled to specific versions of deep learning frameworks like PyTorch. For instance, the latest version of \textit{Signatory} is only compatible with PyTorch 1.9.0, which was released in June 2021. This creates significant versioning and maintenance challenges, limiting the long-term usability of these libraries. Furthermore, the reliance on low-level optimizations, such as fused operations and handwritten backpropagation, complicates the integration of signature computations into modern machine learning pipelines. In this paper, we introduce \textit{Keras Sig}, a pure Python library. Using high-level tensor operations and avoiding low-level primitives, our library achieves competitive performance and is seamless for Python developers. Our key contributions are threefold. First, we present a GPU-optimized implementation of the signature transform that strategically reorganizes computations to maximize parallelism, resulting in substantial speedups compared to existing libraries. Second, we introduce a backend-agnostic design that ensures compatibility with various deep learning frameworks such as Pytorch, Tensorflow, and JAX, addressing versioning challenges and simplifying long-term maintenance. Finally, we provide a user-friendly API that seamlessly integrates with the Keras ecosystem, empowering researchers and practitioners to easily incorporate signature methods into diverse machine learning techniques. The library has already demonstrated its practical value in our concurrent research \cite{inzirillo2024sigkan}, where it successfully replaced a developed TensorFlow wrapper over \textit{iisignature \cite{iisignature}}, leading to enhanced performance. Our experiments demonstrate that \textit{Keras Sig} outperforms existing libraries on GPU hardware, while maintaining comparable performance on CPUs. By eliminating the need for low-level optimizations and framework-specific dependencies, our library provides a robust and future-proof solution for incorporating signature transforms into machine learning workflows.

\section{Related Work}
Introduced in the context of rough path theory \cite{lyons1998}, the signature transform has gained popularity in machine learning as a feature extraction tool for sequential data. \cite{chevyrev2016} and \cite{levin2013} demonstrated the utility of signatures in tasks such as time series classification and handwriting recognition. More recently, the signature transform has been integrated into deep learning architectures, with various applications: from neural differential equations \cite{kidger2020} to generative adversarial networks \cite{ni2020}. Several libraries have been developed to facilitate the computation of signatures in machine learning. \textit{Signatory} proposed by \cite{kidger2021} is the most widely used library, offering CPU and GPU support, as well as handwritten backpropagation for efficient gradient computation. However, \textit{Signatory}'s reliance on C++ code and its tight coupling with specific versions of PyTorch have led to significant versioning issues. For example, the latest version of \textit{Signatory} is only compatible with PyTorch 1.9.0, which is now outdated (torch v2.5.0 available). This limits the library's usability in modern deep learning pipelines, where compatibility with the latest framework versions is crucial. Other libraries, such as \textit{esig} \cite{lyons2017} and \textit{iisignature} \cite{reizenstein2018}, provide CPU-based implementations of the signature transform but lack GPU support and are not optimized for deep learning workflows. Even more recently, \textit{Signax} \cite{anhtong2023} introduced a pure JAX implementation of the signature transform, demonstrating that high-level tensor operations can achieve competitive performance without the need for low-level optimizations. However, \textit{Signax} is limited to the JAX backend and does not provide the same level of framework flexibility as \textit{Keras Sig}. Our work builds on these prior efforts by introducing a backend-agnostic library that combines the performance benefits of GPU-optimized computations with the simplicity and maintainability of high-level tensor operations. By leveraging the Keras 3 framework, \textit{Keras Sig} ensures compatibility with multiple backends (JAX, TensorFlow, and PyTorch), reducing versioning issues and simplifying integration into existing machine learning pipelines. This approach addresses a critical gap in the existing ecosystem, providing a robust and future-proof solution for computing signatures in deep learning applications.

%% file: 2_Innovation.tex
\section{A New View on Signature Computation}
The signature transform, by its mathematical definition, involves computing iterated integrals over a path. For a path $X: [0,T] \to \mathbb{R}^d$, its signature up to depth $N$ is defined as:

\begin{equation}
\text{Sig}^N(X) = \left(1, \int_0^T dX_t, \int_0^T \int_0^{t_2} dX_{t_1} \otimes dX_{t_2}, \dots, \int_0^T \int_0^{t_N} \cdots \int_0^{t_2} dX_{t_1} \otimes \cdots \otimes dX_{t_N}\right)
\end{equation}

In practice, we work with discretized paths, transforming these integrals into finite sums over path increments. Traditional implementations, such as those in \textit{Signatory} \cite{kidger2021}, compute these sums sequentially by iterating over path increments and applying fused operations at each step. This approach, while memory-efficient and well-suited for CPU execution, fails to fully exploit the parallel processing capabilities of modern GPU architectures. In this section, we present a novel reformulation of signature computation that fundamentally changes how these iterated sums are evaluated. Our approach reorganizes the computation to maximize parallel operations while maintaining mathematical equivalence with the traditional sequential method. Instead of processing path increments sequentially, we leverage the associative property of tensor operations to express each degree of the signature as a combination of parallel matrix operations and efficient cumulative sums. This reformulation is particularly well-suited for GPU execution, where parallel operations on large tensors can be performed with high efficiency.
\subsection{Traditional Approach: Sequential Fused Operations}
The traditional approach computes signatures sequentially along the path. For a path $X = (X_1, X_2, \dots, X_L) \in \mathbb{R}^{L \times d}$, with increments $\Delta X_i = X_{i+1} - X_i$, we compute signatures for each prefix of the path using an iterative process.

First, for any increment $\Delta X$, we compute its restricted exponential up to depth $N$:
\begin{equation}
\exp^N(\Delta X) = \left(1, \Delta X, \frac{\Delta X \otimes \Delta X}{2!}, \dots, \frac{\Delta X^{\otimes n}}{n!}, \dots, \frac{\Delta X^{\otimes N}}{N!}\right)
\end{equation}

Then, for the path prefix up to index $k$, we compute recursively:
\begin{equation}
\begin{aligned}
\text{Sig}^N_1 &= \exp^N(\Delta X_1) \\
\text{Sig}^N_k &= \text{Sig}^N_{k-1} \boxtimes \exp^N(\Delta X_k) \quad \text{for } k = 2,\dots,L
\end{aligned}
\end{equation}

where $\boxtimes$ denotes the multiplication of tensors defined as:
\begin{equation}
\begin{aligned}
(a_1, \dots, a_N) \boxtimes (b_1, \dots, b_N) &= (c_1, \dots, c_N) \\
\text{with } c_n &= \sum_{i+j=n} a_i \otimes b_j
\end{aligned}
\end{equation}

For a batch of paths $X \in \mathbb{R}^{B \times L \times d}$, this becomes:
\begin{equation}
\begin{aligned}
\text{Sig}^N_{b,1} &= \exp^N(\Delta X_{b,1}) \\
\text{Sig}^N_{b,k} &= \text{Sig}^N_{b,k-1} \boxtimes \exp^N(\Delta X_{b,k}) \quad \text{for } k = 2,\dots,L
\end{aligned}
\end{equation}
where $b$ indexes the batch dimension. An important property of this approach is that computing $\text{Sig}^N_{k}$ only requires access to $\text{Sig}^N_{k-1}$ and $\Delta X_k$, allowing memory-efficient processing of long sequences. However, this sequential computation pattern, requiring $L-1$ iterations of tensor products for each path, becomes a performance bottleneck on GPU architectures designed for parallel operations.

\subsection{GPU-Optimized Approach: Parallel Sequence Processing}
Our approach reorganizes the signature computation to maximize parallel operations across the sequence dimension. For a path $X \in \mathbb{R}^{B \times L \times d}$, we compute signatures of different depths simultaneously using cumulative sums and parallel tensor operations. Key to our approach is the observation that the $n$-th degree term of the signature can be expressed as:
\begin{equation}
\text{Sig}^n(X) = \sum_{1 \leq i_1 < \dots < i_n \leq L} \Delta X_{i_1} \otimes \dots \otimes \Delta X_{i_n}
\end{equation}

This can be computed efficiently using cumulative sums and pre-divided increments. For each degree $n$, we define:
\begin{equation}
\Delta X^{(n)} = \frac{\Delta X}{n!}
\end{equation}

Our parallel computation then proceeds as follows:
\begin{equation}
\begin{aligned}
    S_{1,i} &= \sum_{j=1}^{i} \Delta X_j\\
S_{2,i} &= \sum_{j=1}^{i} \left( \Delta X_j \otimes S_{1,j} \right)\\
S_{3,i} &= \sum_{j=1}^{i} \left( \Delta X_j \otimes \left( S_{2,j} + \Delta X^{(2)} \otimes S_{1,j} \right) \right)\\
S_{4,i} &= \sum_{j=1}^{i} \left( \Delta X_j \otimes \left( S_{3,j} + \Delta X^{(2)} \otimes S_{2,j} + \Delta X^{(3)} \otimes S_{1,j} \right) \right)\\
&\vdots
\end{aligned}
\end{equation}

where $\text{cumsum}$ operates along the sequence dimension and $\otimes$ represents batched tensor product operations preserving both batch and sequence dimensions:
\begin{equation}
(\Delta X \otimes S)_{b,k} = \Delta X_{b,k} \otimes S_{b,k} \quad \text{for all } b,k
\end{equation}

For a given prefix length $k$, the signature up to depth $N$ is then obtained by concatenating:
\begin{equation}
\text{Sig}^N_{b,k} = [S_{1,b,k}, S_{2,b,k}, \dots, S_{N,b,k}]
\end{equation}

This formulation replaces sequential operations with parallel matrix multiplications and cumulative sums, operations highly optimized on GPU architectures. While this approach requires more memory to store intermediate results, it significantly reduces the number of sequential operations required, leading to substantial performance improvements on GPU hardware.

\subsection{Advantages of the GPU-Optimized Approach}
Our approach offers several advantages over traditional methods:
\begin{itemize}
    \item \textbf{Optimized Sequential Operations}: While our method still requires one sequential operation—the cumulative sum along the sequence dimension—this operation is highly optimized in modern GPU architectures compared to the more complex fused tensor operations used in traditional approaches.
    
    \item \textbf{Parallel Processing}: By reorganizing the computation to parallelize operations across the sequence dimension, our method leverages GPU's strength in performing large matrix operations simultaneously. The only sequential dependency occurs in the cumulative sum, which is implemented efficiently using parallel scan algorithms.
    
    \item \textbf{Efficient Memory Access}: Although our approach requires more memory to store intermediate results, it benefits from GPU's high memory bandwidth and coalesced memory access patterns, as most operations are performed on contiguous blocks of memory.
    
    \item \textbf{Scalability}: The use of batched tensor operations allows efficient processing of multiple sequences simultaneously, making our method particularly effective for the large batch sizes common in deep learning applications.
\end{itemize}

\subsection{Automatic Backend Selection}
Our library automatically selects the most appropriate implementation based on the available hardware. On GPUs, it employs the parallel approach described above, leveraging the hardware's capability for efficient cumulative sum operations and parallel matrix computations. On CPUs, where the memory access patterns and parallel processing capabilities differ significantly, the library switches to an implementation following the traditional sequential approach used by \textit{Signatory} and \textit{Signax}. This automatic selection ensures optimal performance across different hardware configurations without requiring user intervention.

\subsection{Limitations and Trade-offs}
While our GPU-optimized approach significantly improves performance on GPUs, it comes with certain trade-offs:
\begin{itemize}
    \item \textbf{Memory Requirements}: The parallel nature of our approach necessitates storing intermediate results for the entire sequence, leading to higher memory usage compared to sequential methods. This can become a limiting factor for very long sequences or high-dimensional data.
    
    \item \textbf{Hardware Specificity}: The performance benefits are specifically tied to GPU architectures with efficient parallel scan operations and high memory bandwidth. On CPUs or older GPU architectures with less efficient cumulative sum implementations, the traditional sequential approach may be more appropriate.
\end{itemize}

In the following section, we present experimental results that quantify these trade-offs and demonstrate the substantial performance benefits our approach achieves on modern GPU hardware.

%% file: 3_Performance_Analysis.tex
\section{Performance Analysis}
To assess the improvements in our path signature computation approach, we conducted two distinct tests. The first test compares the computation time required by different libraries to process signature batches, varying both sequence length and signature degree. 
Since this package is primarily designed for integration with Keras models, our second test embeds the signature computation within deep learning models, incorporating learnable weights both before and after the signature layer. While our primary innovation focuses on GPU optimization, we performed tests on both GPU and CPU architectures to demonstrate our approach's versatility. Though optimized for GPU computation, our implementation automatically falls back to a CPU implementation when no GPU is detected, using a method similar to Signax but implemented with Keras operations instead of JAX, ensuring compatibility across different hardware configurations. One of Keras 3's key strengths is its ability to operate with multiple backends, so we tested our package across all supported backends. However, it's important to note that while our package runs on any backend, the underlying implementation differences create certain constraints. For instance, models using Keras Sig with PyTorch backend cannot be JIT compiled (a limitation also present in Signatory), whereas the JAX backend can take full advantage of XLA optimizations.To ensure consistent and comparable results, all experiments were conducted on machines rented from \textit{vast.ai}, equipped with a Ryzen 5900X CPU and an RTX 4090 GPU. We established specific software environments for each library under test:
\begin{itemize}
    \item \textit{Signatory}: Implemented using Docker image\\
        \texttt{pytorch/pytorch:1.9.0-cuda11.1-cudnn8-runtime},\\
        running \textit{Signatory} version 1.2.6.1.9.0.
    \item \textit{TensorFlow}: Deployed using Docker image\\
        \texttt{tensorflow/tensorflow:2.16.1-gpu}.
    \item \textit{Keras Sig} (JAX and PyTorch backends): Built on Docker image\\
        \texttt{nvidia/cuda:12.6.0-runtime-ubuntu22.04}, utilizing\\
        \textit{Keras} version 3.7.0, \textit{Signax} version 0.2.1, \textit{Keras Sig} version 1.0.2,\\
        \textit{JAX} version 0.4.38, and \textit{PyTorch} version 2.5.1+cu124.
\end{itemize}
In addition to deep learning-specific libraries, we evaluated standard signature computation packages: esig (version 1.0.0) and iisignature (version 0.24). For the model training evaluation, we also included iisignature\_tensorflow\_2 (version 0.1.0), a TensorFlow 2 wrapper built around iisignature.

\subsection{Direct Signature Computation}
We evaluated computation times across different batch sizes, sequence lengths, and signature depths, comparing our library with existing implementations. Our results demonstrate the efficiency and scalability of our approach, with particularly strong performance on GPU hardware.

\subsubsection{GPU Performance}

\begin{figure}[ht]
    \centering
    \includegraphics[width=0.6\columnwidth]{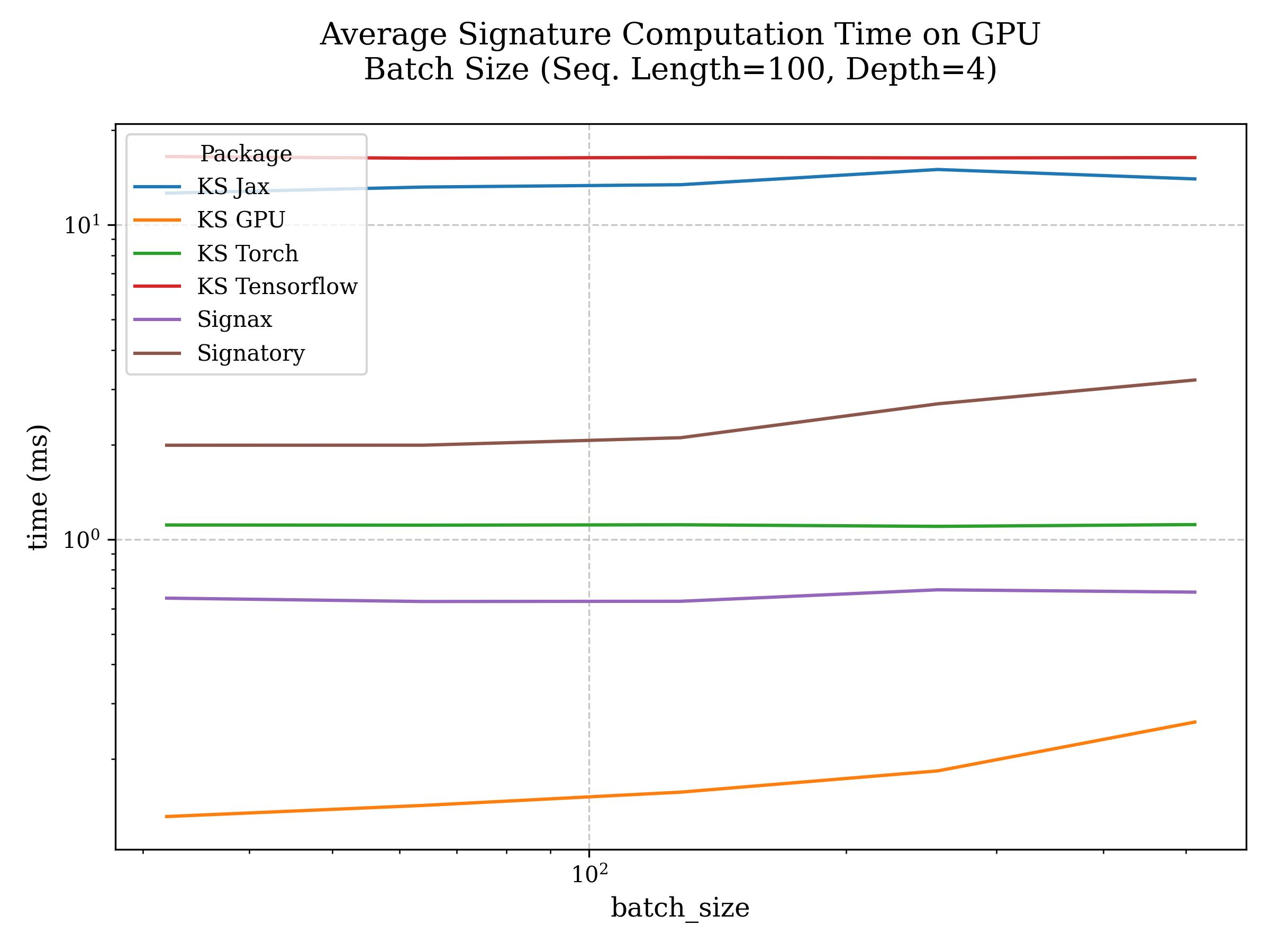}
    \caption{Average signature computation time on GPU as a function of batch size (Seq. Length=100, Depth=4).}
    \label{fig:gpu_batch_size}
\end{figure}

\begin{figure}[ht]
    \centering
    \includegraphics[width=0.6\columnwidth]{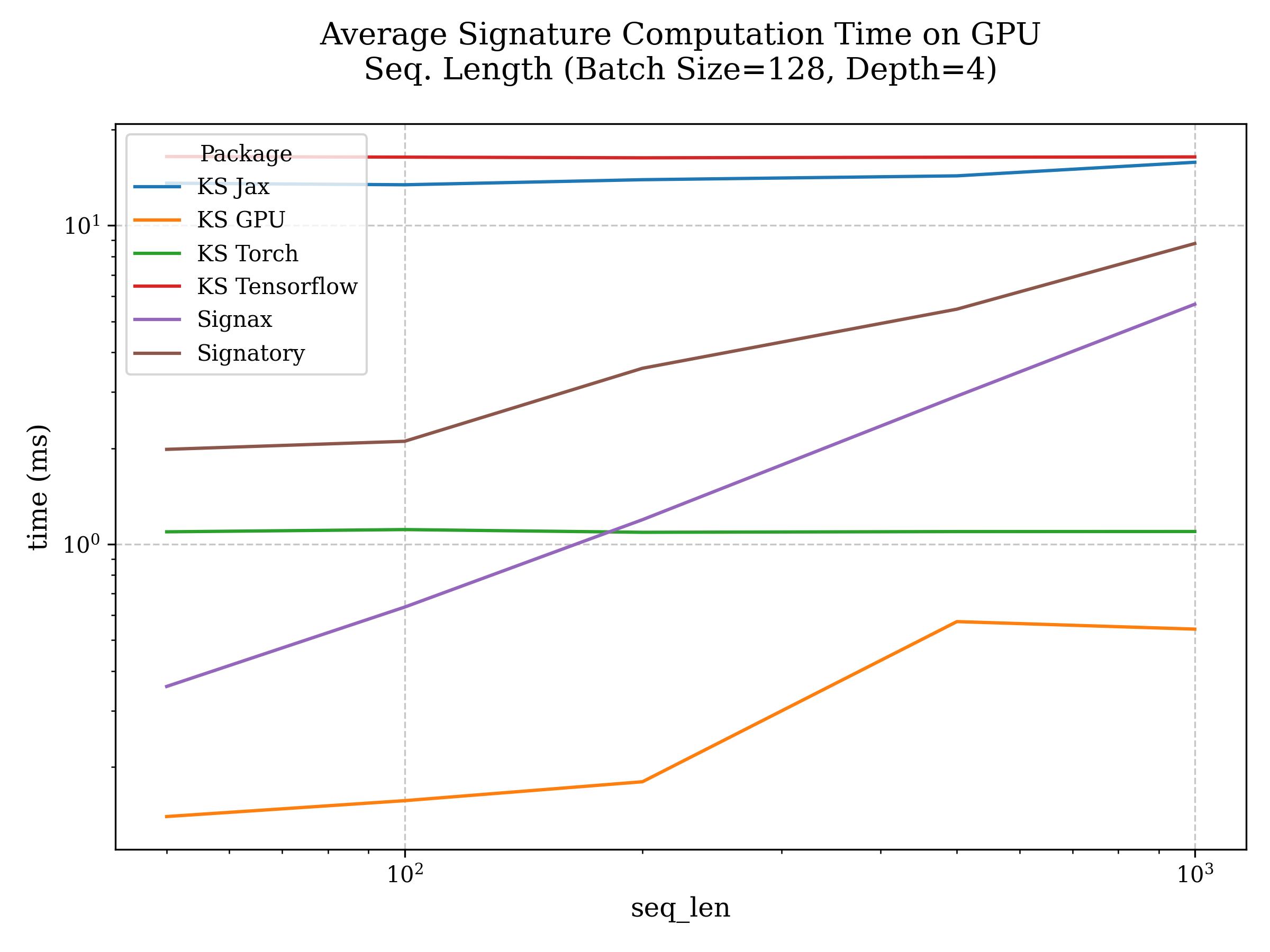}
    \caption{Average signature computation time on GPU as a function of sequence length (Batch Size=128, Depth=4).}
    \label{fig:gpu_seq_len}
\end{figure}

Table \ref{tab:sig_computation_gpu} presents signature computation times on GPU across various configurations. For brevity, \textit{Keras Sig} is abbreviated as KS in the following tables. It's important to note that KS GPU refers to our pure JAX implementation of the GPU-optimized function, while other KS entries represent implementations using standard Keras functions.
\input{tables/signature_computation_gpu}
The first notable observation is the significant variation in Keras Sig's performance across different backends. This variation stems from Keras's architecture, which is primarily designed for training deep learning models rather than pure scientific computation. When using JAX or TensorFlow backends, Keras compiles the function for each execution without caching the compilation, necessitating costly recompilation for every call and resulting in suboptimal performance.
However, with the PyTorch backend, where this compilation behavior differs, our GPU implementation (which is automatically detected) demonstrates significant performance improvements over Signatory, despite both using PyTorch as their foundation. This improvement, reducing computation time by a factor of 2 to 3, provides initial validation of our approach's effectiveness. Furthermore, Signax, which employs computation methods similar to Signatory, achieves remarkable performance gains through its use of JAX, delivering 3 to 6 times faster computation times. This highlights how modern computational libraries like JAX, despite their high-level abstraction, can leverage XLA compilation to outperform lower-level C++ implementations while maintaining more readable and maintainable code.
To fully appreciate the impact of our GPU optimization, we can directly compare KS GPU with Signax, as both are implemented in pure JAX. When GPU hardware is available, the performance gains are substantial, with our approach showing 5 to 10-fold improvements over Signax, which translates to 10 to 20-fold improvements over Signatory.To better understand how different parameters affect performance, we created several visualizations analyzing key metrics. Figure \ref{fig:gpu_batch_size} illustrates the average signature computation time on GPU as a function of batch size, with sequence length fixed at 100 and depth at 4. While \textit{Keras Sig} GPU demonstrates the fastest computation times across all batch sizes, its performance curve shows a steeper slope compared to other implementations. This behavior can be attributed to different parallelization strategies: Signax primarily parallelizes across batch dimensions, meaning its efficiency increases with batch size as long as memory constraints are not exceeded. In contrast, Keras Sig may encounter memory congestion at larger batch sizes. However, this limitation is less significant in practice, as deep learning models typically perform better with moderate batch sizes, even though larger batches might offer faster training times.
Figure \ref{fig:gpu_seq_len} demonstrates how sequence length affects computation time, with batch size fixed at 128 and depth at 4. \textit{Keras Sig} exhibits a markedly different behavior compared to its batch size performance. While Signatory and Signax show linear increases in computation time as sequence length grows, our GPU approach maintains nearly constant computation time across different sequence lengths, with only a single notable increase at a specific threshold—likely due to matrix computation constraints. This pattern validates our innovative approach of processing the entire sequence as a single matrix multiplication operation, resulting in computation times that remain largely independent of sequence length, provided the data fits within available memory.

\begin{figure}[ht]
    \centering
    \includegraphics[width=0.6\columnwidth]{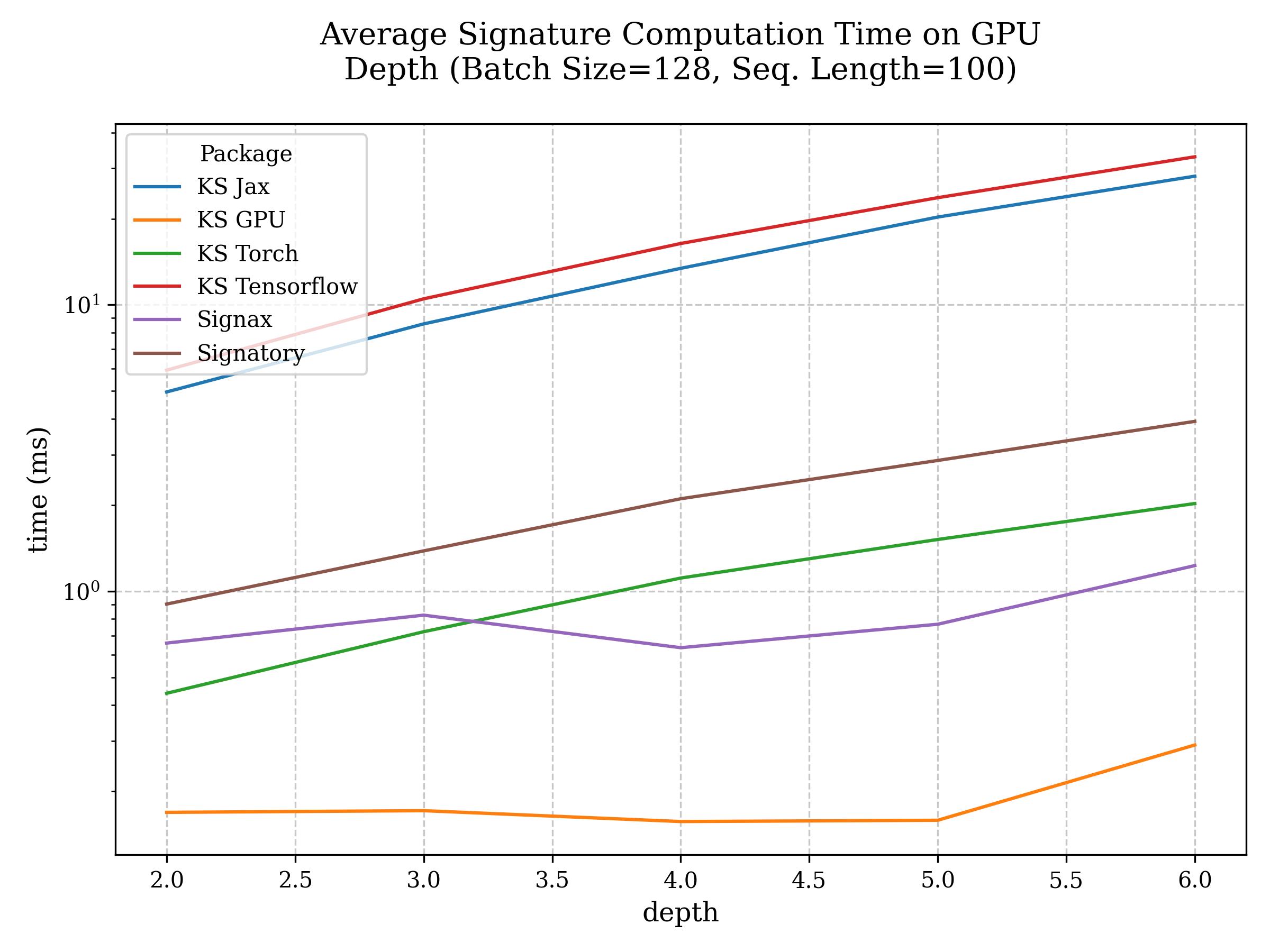}
    \caption{Average signature computation time on GPU as a function of signature depth (Batch Size=128, Seq. Length=100).}
    \label{fig:gpu_depth}
\end{figure}

Figure \ref{fig:gpu_depth} examines the relationship between signature depth and computation time, with batch size fixed at 128 and sequence length at 100. The results present a more complex pattern than initially anticipated, deviating from expected uniform trends across implementations. For lower signature degrees, Signatory exhibits a nearly linear increase in computation time, while Signax and our GPU approach show more gradual growth. At higher degrees, all implementations demonstrate increased computation times, though JAX-based implementations maintain a performance advantage, likely due to JAX's superior optimization capabilities for operations with fewer degrees

\subsubsection{CPU Performance}

Table \ref{tab:sig_computation_cpu} presents signature computation times across different configurations on CPU hardware. For this analysis, we expanded our comparison to include standard signature libraries—iisignature and esig—which were excluded from GPU testing due to their lack of GPU support.The results show similar backend-related effects as observed in GPU testing, with JAX and TensorFlow backends exhibiting performance limitations that make them suboptimal for CPU computation. Notably, Signax and Signatory demonstrate comparable performance levels on CPU, with their relative efficiency varying by use case due to backend-specific optimizations. Two significant observations emerge regarding Keras Sig: First, the PyTorch backend implementation performs slightly worse than Signatory, a consequence of additional abstraction layers introduced to maintain backend agnosticism. Second, and more notably, our JAX GPU-optimized version shows reduced performance compared to Signax on CPU—a reversal of our GPU results. This performance inversion clearly demonstrates how our optimization strategy's effectiveness is hardware-dependent.

\begin{figure}[ht]
    \centering
    \includegraphics[width=0.6\columnwidth]{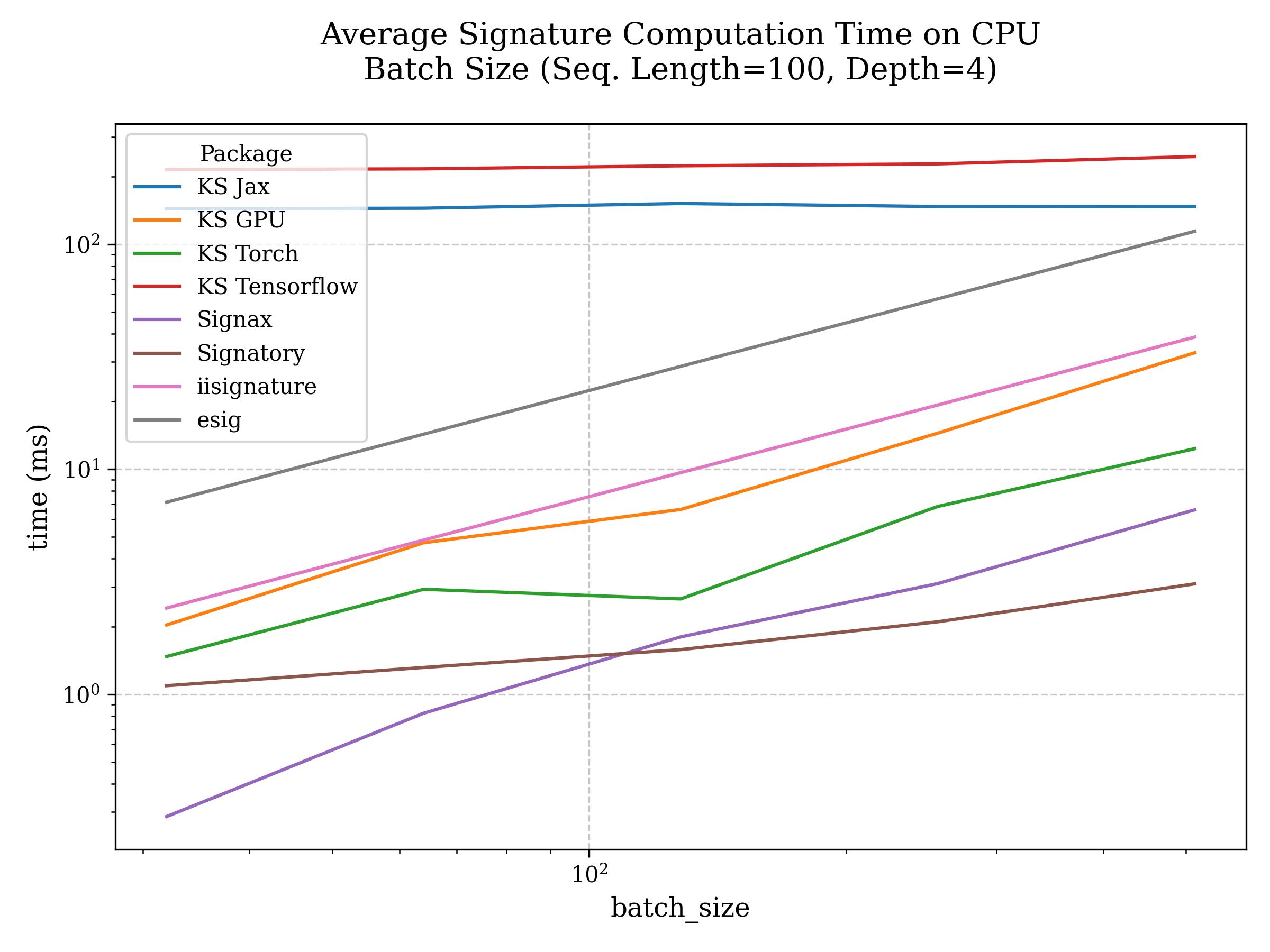}
    \caption{Average signature computation time on CPU as a function of batch size (Seq. Length=100, Depth=4).}
    \label{fig:cpu_batch_size}
\end{figure}

\input{tables/signature_computation_cpu}
Figure \ref{fig:cpu_batch_size} presents the average signature computation time on CPU as a function of batch size, with sequence length fixed at 100 and depth at 4. The results reveal Signatory's superior handling of increasing batch sizes on CPU architecture. This advantage likely stems from the fundamental differences in parallelization strategies: while Signax employs batch-level vectorization—highly effective on GPUs but less impactful on CPUs—Signatory's native parallelism management appears better suited for CPU execution.

\begin{figure}[ht]
    \centering
    \includegraphics[width=0.6\columnwidth]{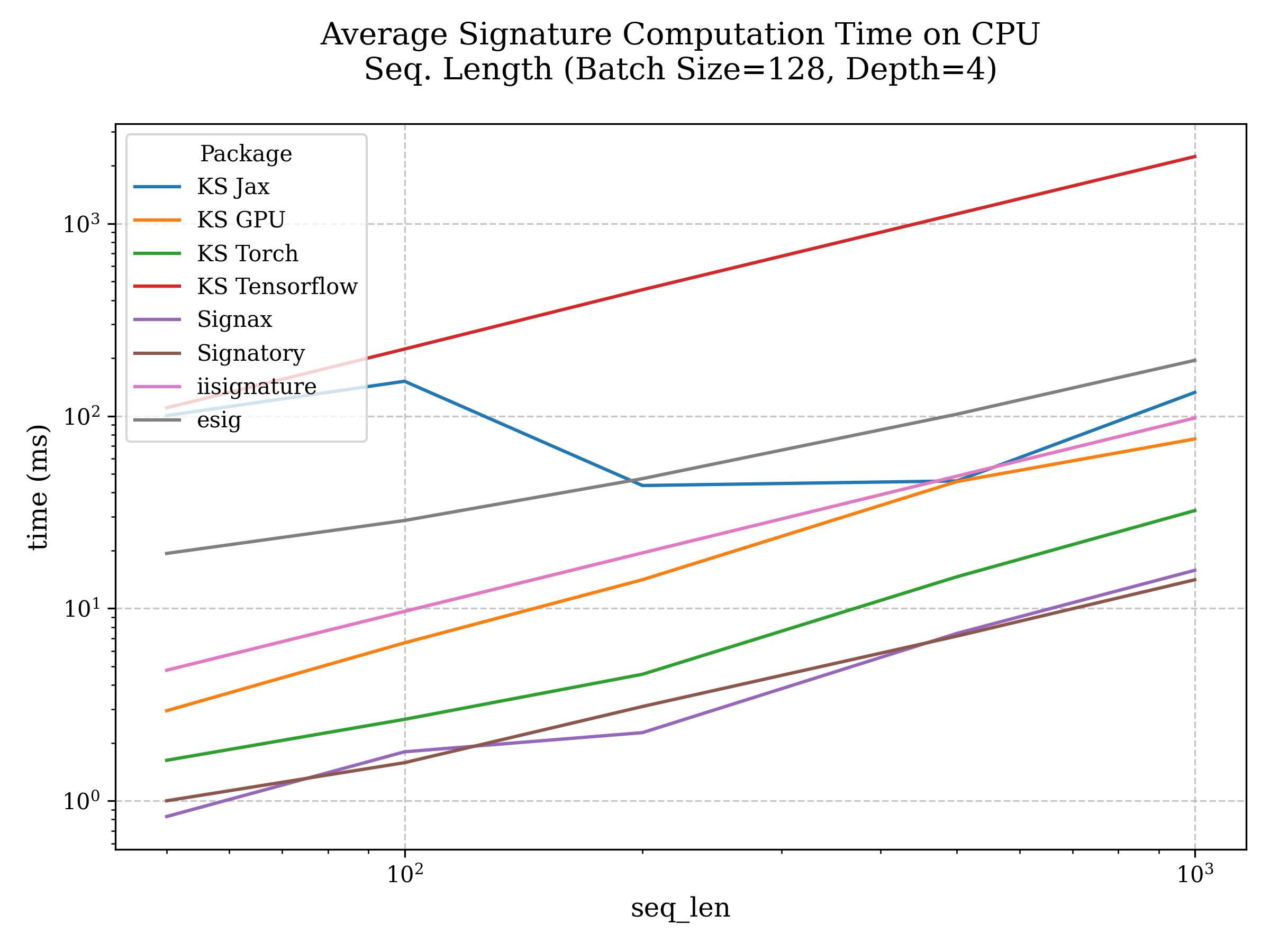}
    \caption{Average signature computation time on CPU as a function of sequence length (Batch Size=128, Depth=4).}
    \label{fig:cpu_seq_len}
\end{figure}

Figure \ref{fig:cpu_seq_len} demonstrates the relationship between sequence length and computation time, with batch size fixed at 128 and depth at 4. On CPU, \textit{Keras Sig} loses its distinctive advantage observed in GPU testing, instead exhibiting a linear increase in computation time similar to other packages. This performance shift occurs because the CPU architecture cannot support the simultaneous matrix multiplications that enable our GPU optimization's efficiency.

\begin{figure}[ht]
    \centering
    \includegraphics[width=0.6\columnwidth]{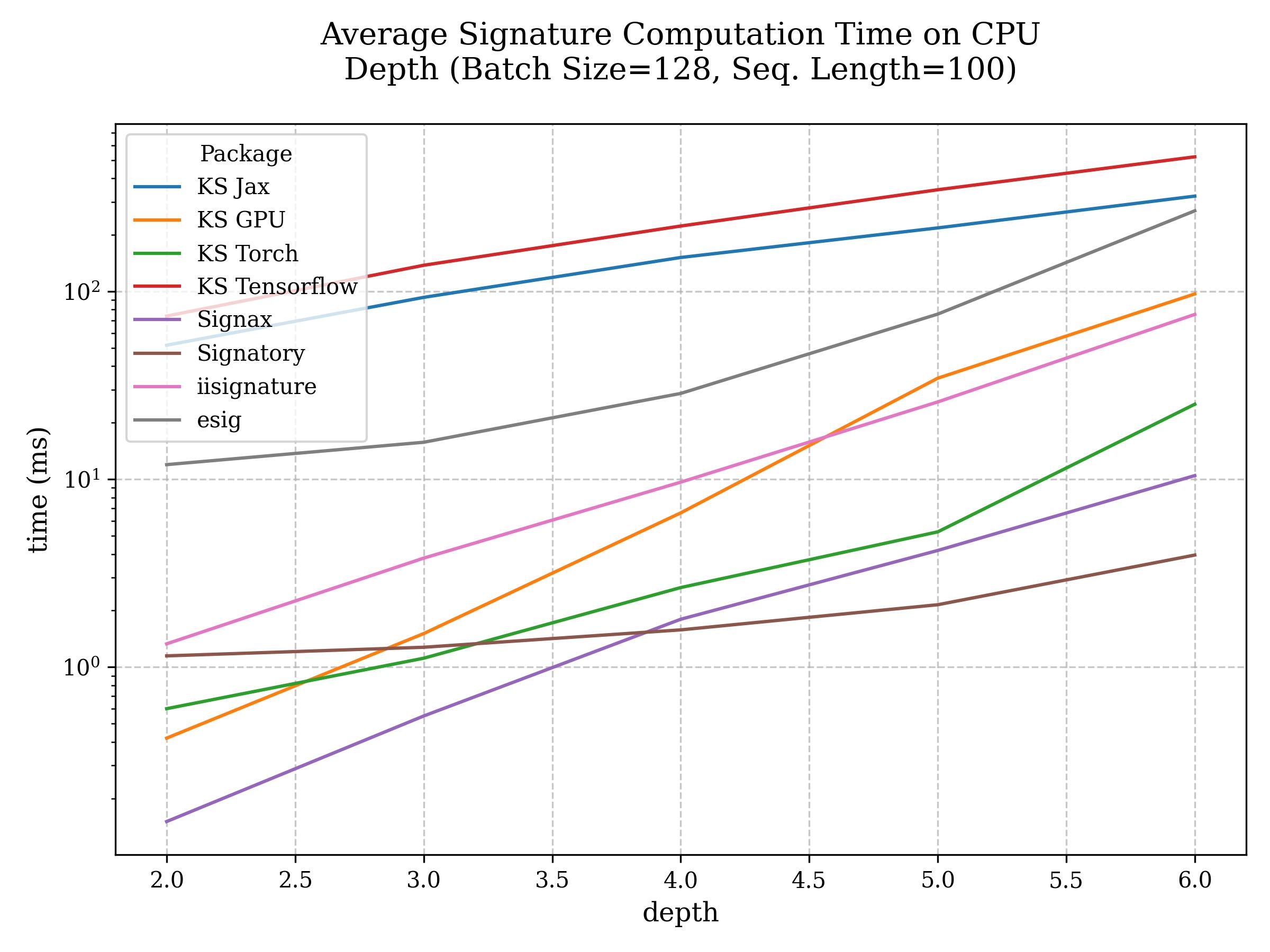}
    \caption{Average signature computation time on CPU as a function of signature depth (Batch Size=128, Seq. Length=100).}
    \label{fig:cpu_depth}
\end{figure}

Figure \ref{fig:cpu_depth} examines how signature depth affects computation time, with batch size fixed at 128 and sequence length at 100. The results largely mirror those observed in the sequence length analysis, with one notable distinction: Signatory demonstrates superior performance when handling increased signature depths, suggesting its implementation is particularly well-optimized for managing higher-order signatures on CPU architecture.

In conclusion, our analysis demonstrates that our proposed GPU optimization significantly improves signature computation performance, though its effectiveness is highly hardware-dependent. For CPU-based computations, Signax emerges as the most practical and efficient solution, offering both easier installation compared to Signatory and freedom from Keras's compilation management overhead.

\subsection{Signature in deep neural networks}

As Keras Sig is specifically designed for integrating signatures into deep learning models, this section presents our most significant results.

To thoroughly evaluate both forward and backward computation, we designed a simple yet representative model with trainable layers both before and after the signature computation. We implemented this architecture using Keras for Keras Sig and Signax (the latter exclusively with JAX backend), and PyTorch for Signatory. The model architecture can be formally described as follows. Given an input sequence \(X \in \mathbb{R}^{B\times L\times 20}\), where \(B\) is the batch size and \(L\) is the sequence length, the model applies these transformations:

\begin{equation}
\begin{aligned}
& \text{Input: } & X &\in \mathbb{R}^{B\times L\times 20} \\[0.5em]
& \text{Layer 1: } & Z &= \phi(XW_1 + b_1), & W_1 &\in \mathbb{R}^{20\times d}, & b_1 &\in \mathbb{R}^d,\\[0.5em]
& \text{Signature: } & S &= \text{Sig}(Z), & Z &\in \mathbb{R}^{B\times L\times d}, & S &\in \mathbb{R}^{B\times D}, \\[0.5em]
& \text{Layer 2: } & Y &= SW_2 + b_2, & W_2 &\in \mathbb{R}^{D\times 10}, & b_2 &\in \mathbb{R}^{10},
\end{aligned}
\end{equation}
where,$d$ represents the signature input size (ranging from 2 to 10) and $D$ is the signature output dimension determined by $d$ and the signature depth. The final output $Y$ produces 10-dimensional vectors for each element in the batch. This architecture provides a clear framework for evaluating the impact of signature computation within a neural network context. For benchmarking purposes, we generated synthetic data consisting of input tensors \(X \in \mathbb{R}^{12765 \times L \times 20}\) and target tensors $Y \in \mathbb{R}^{12765 \times 10}$, where \(L\) represents the sequence length. The training process used a batch size of 128, resulting in 99 complete batches and one final batch of 65 samples. This deliberate choice of an incomplete final batch helped verify the signature layer's robustness in handling varying batch sizes during training. The performance assessment consisted of training each model configuration for 10 epochs across different backends, measuring the total training time for completion.

\subsubsection{GPU Performance}

Table \ref{tab:sig_computation_gpu} shows model training times on GPU across various configurations of signature input size, sequence length, and signature degree.
\input{tables/model_training_gpu}
In contrast to direct signature computation, where Keras's compilation management posed challenges, all Keras Signature configurations demonstrate viable performance in the training context. For shorter sequence lengths, Keras Sig with JAX backend achieves training times comparable to the Signax package. However, as sequence length increases, our implementation demonstrates significant improvements, reducing training times by up to 55\% compared to Signax, which itself already outperforms Signatory by a factor of 2.However, these performance gains come with certain limitations, particularly regarding signature output dimensionality. When training models with larger signatures (e.g., input size of 10 and depth of 4), the performance gap between methods narrows significantly. This convergence can be attributed to memory constraints: for instance, with these parameters, the signature output dimension reaches 11,110 values. During computation, this must be maintained across the full sequence for the entire batch, resulting in substantial memory requirements—for a sequence length of 500 and batch size of 128, the intermediate computations involve more than 700 million values. Examining Signatory's performance, we observe that it consistently runs approximately twice as slow as its Signax counterpart across all tested configurations, demonstrating JAX's XLA compilation advantages over traditionally optimized C++ code. A crucial distinction emerges in model compilation capabilities: while models using Signax or Keras Sig can be compiled, PyTorch models incorporating Signatory cannot. Though this limitation appears minor in our simple benchmark model with its modest 2x performance penalty, the impact becomes more significant in complex architectures like transformers, where signature computation represents only a fraction of the total computational load. In such cases, the inability to compile the entire model could lead to substantial performance degradation. Regarding other backends, while they show slightly reduced performance compared to JAX, they remain competitive with Signax's execution times. This is in line  with our expectations, given JAX's more recent development and optimization-focused design. The PyTorch backend implementation of Keras Sig outperforms Signatory, confirming that our GPU optimization strategy translates effectively across different backends. TensorFlow presents an interesting case: it slightly outperforms the PyTorch backend for small signature sizes but exhibits exponential performance degradation with larger signatures. While this behavior likely stems from TensorFlow's internal architecture, the precise mechanism remains unclear and warrants further investigation.

To provide deeper insight into performance characteristics, we developed visualizations analyzing the impact of various parameters.

\begin{figure}[ht]
    \centering
    \includegraphics[width=0.6\columnwidth]{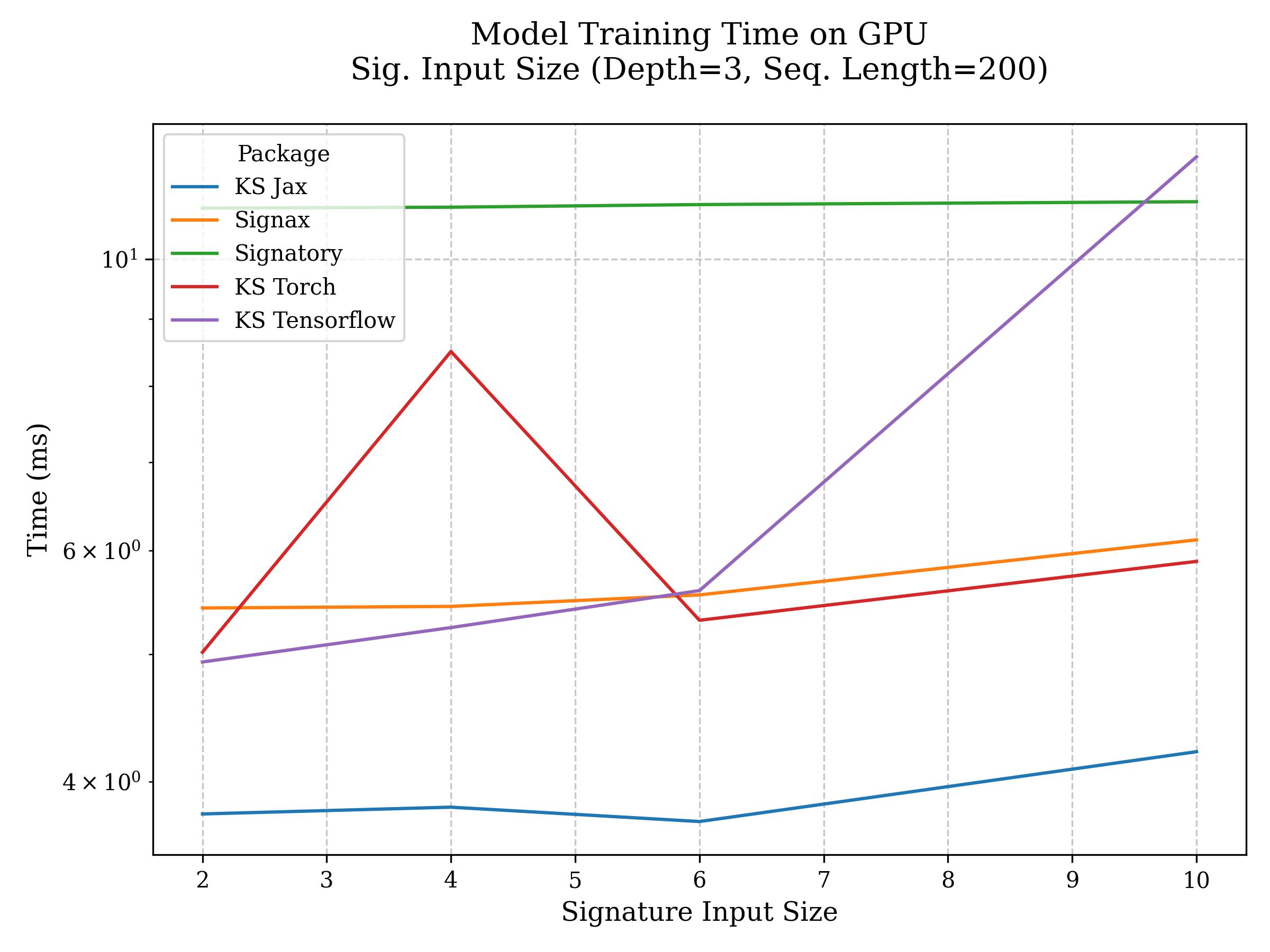}
    \caption{Model Trainig time on GPU as a function of Sig. Input Size (Seq. Length=200, Depth=3).}
    \label{fig:gpu_sig_size_train}
\end{figure}

Figure \ref{fig:gpu_sig_size_train} illustrates training time as a function of signature input size, with sequence length fixed at 200 and signature degree at 3. The visualization reveals distinct performance patterns across backends and implementations. Most approaches demonstrate similar scaling behavior with respect to signature input size, with TensorFlow being a notable exception, showing significant performance degradation at larger signature dimensions. Notably, our PyTorch backend implementation, despite PyTorch's generally lower performance compared to JAX, achieves comparable speeds to the JAX-based Signax implementation. This equivalence demonstrates the effectiveness of our optimization strategy, which successfully compensates for underlying backend performance differences.

\begin{figure}[ht]
    \centering
    \includegraphics[width=0.6\columnwidth]{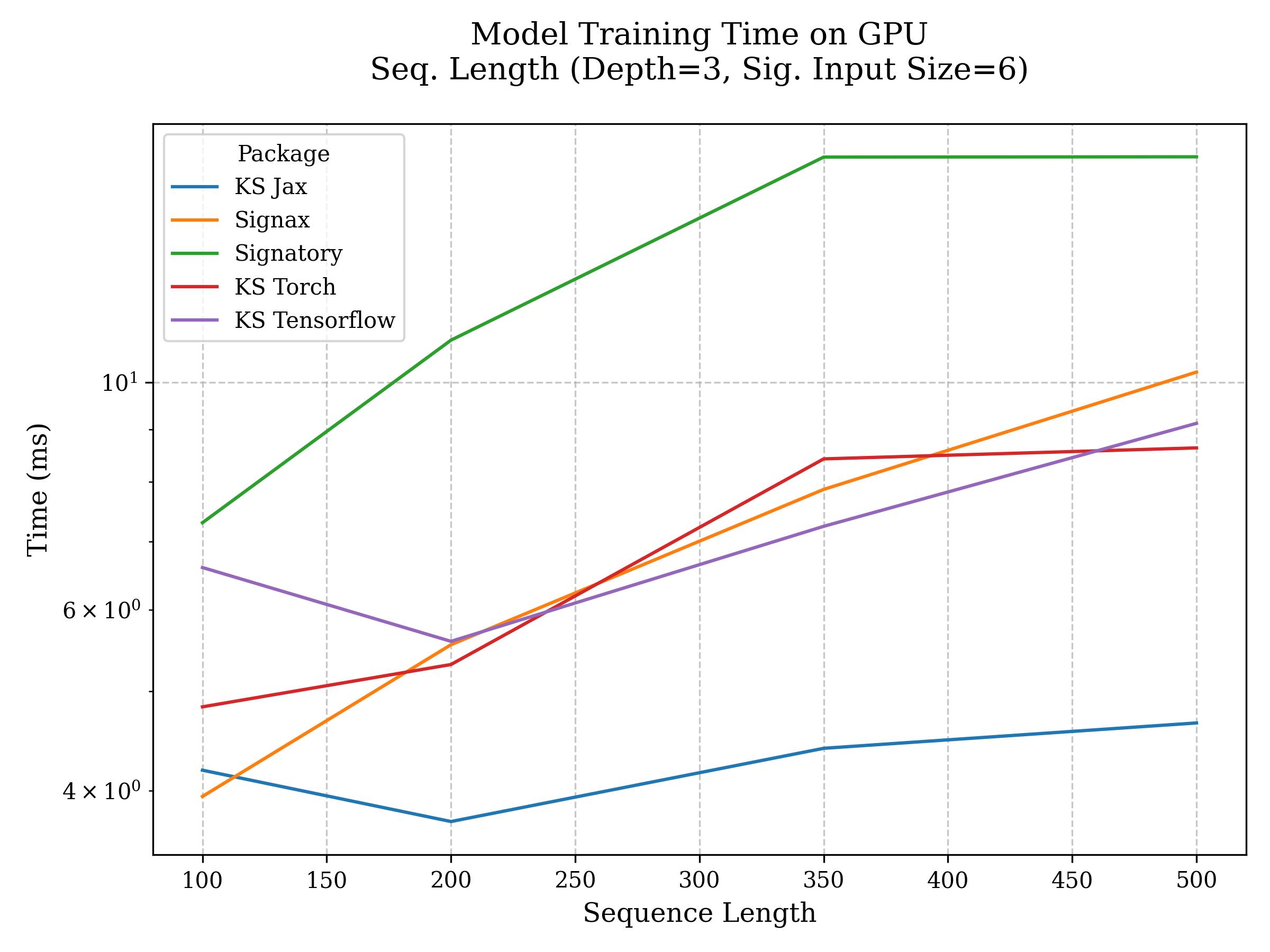}
    \caption{Model Trainig time on GPU as a function of Seq. Length (Sig. Input Size=6, Depth=3).}
    \label{fig:gpu_seqlen_train}
\end{figure}

Figure \ref{fig:gpu_seqlen_train} examines the relationship between sequence length and training time, with signature input size fixed at 6 and degree at 3. The results validate our earlier findings from direct signature computation: our approach demonstrates significantly better scaling with sequence length compared to both Signax and Signatory. This improved scaling behavior can be attributed to our optimized matrix multiplication strategy, which processes sequences more efficiently than traditional iterative approaches.

\begin{figure}[ht]
    \centering
    \includegraphics[width=0.6\columnwidth]{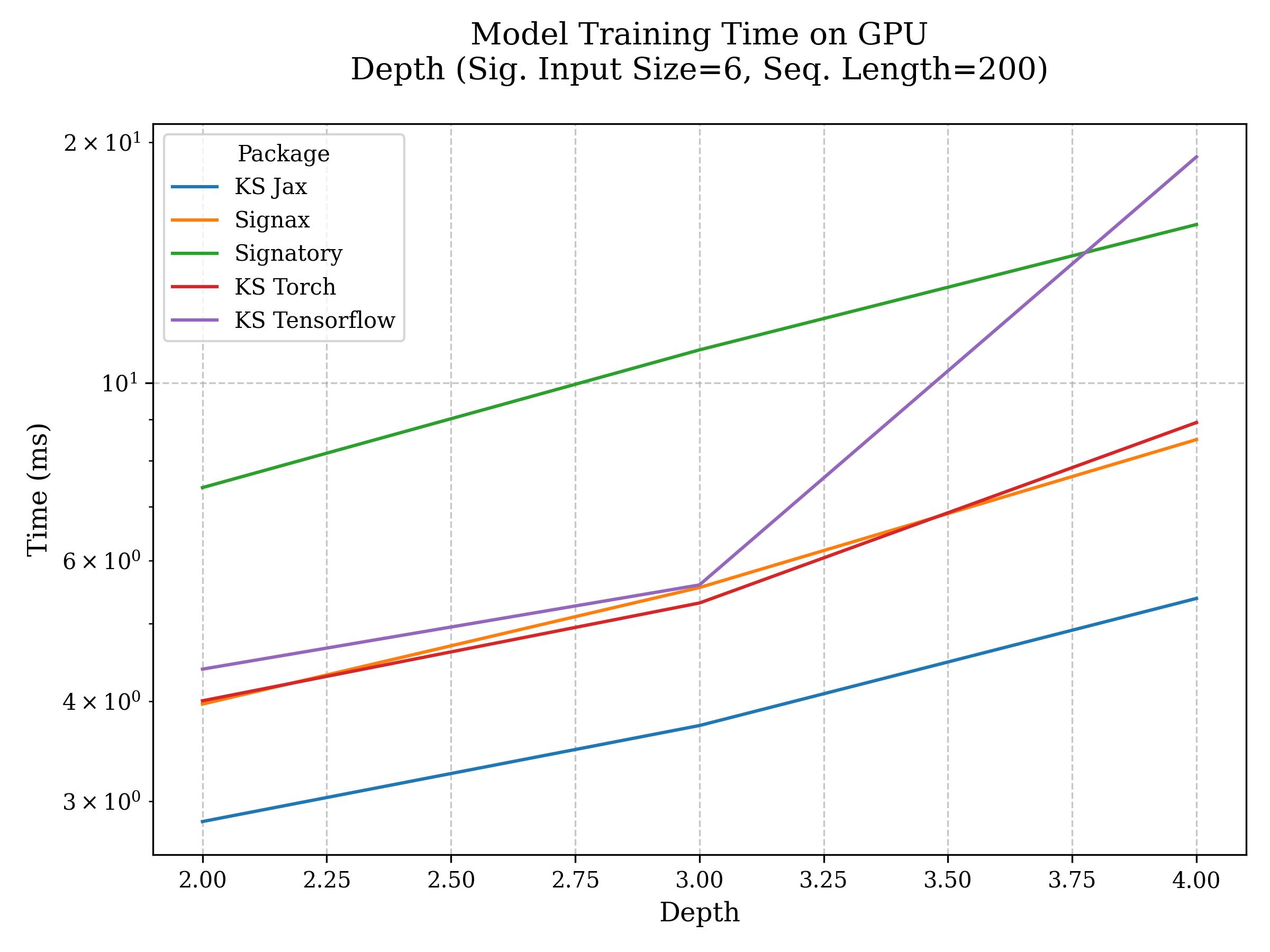}
    \caption{Model Trainig time on GPU as a function of Depth (Sig. Input Size=6, Seq. Length=200).}
    \label{fig:gpu_depth_train}
\end{figure}

Figure \ref{fig:gpu_depth_train} analyzes the impact of signature degree on training time, with signature input size fixed at 6 and sequence length at 200. A notable observation is that for moderate signature input sizes, all implementations except TensorFlow exhibit similar scaling behavior with respect to signature degree. This uniform scaling suggests that signature degree alone does not account for the performance differences between Signax and our approach. Rather, the performance divergence emerges from the interaction between signature input size and degree, becoming particularly pronounced when their combination approaches memory capacity limits.

\subsubsection{CPU Performance}

\begin{figure}[ht]
    \centering
    \includegraphics[width=0.6\columnwidth]{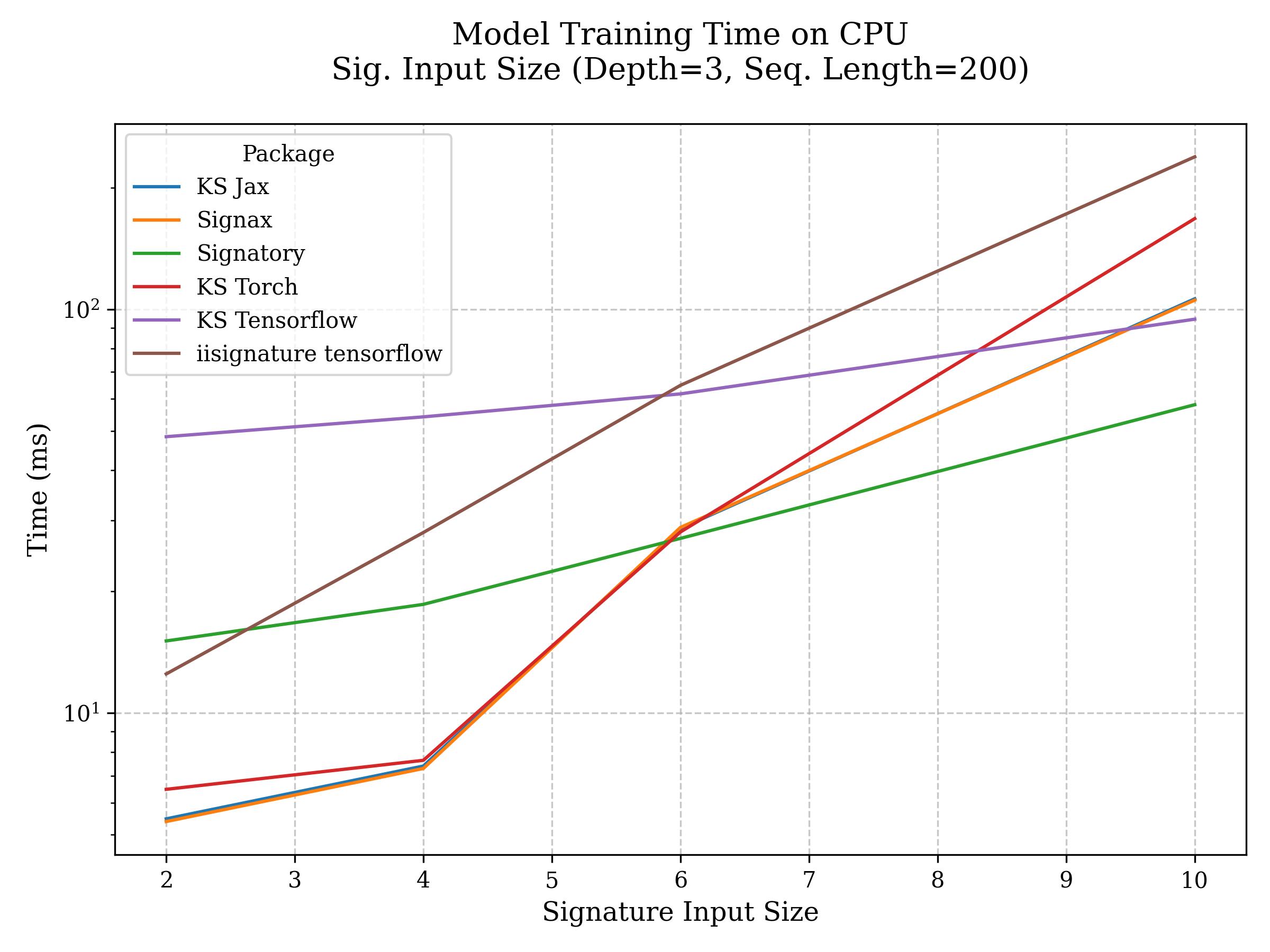}
    \caption{Model Trainig time on CPU as a function of Sig. Input Size (Seq. Length=200, Depth=3).}
    \label{fig:cpu_sig_size_train}
\end{figure}

\begin{figure}[ht]
    \centering
    \includegraphics[width=0.6\columnwidth]{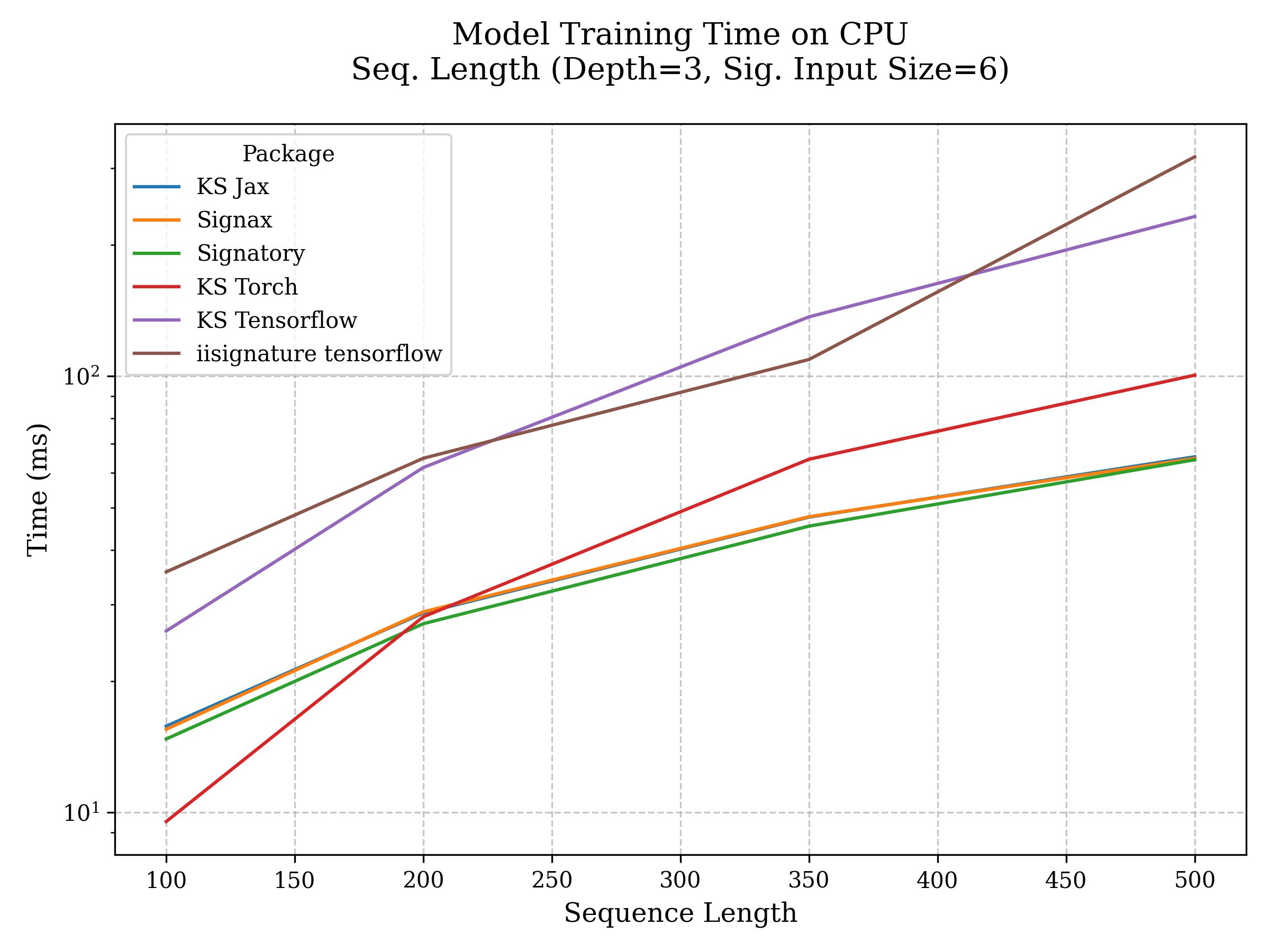}
    \caption{Model Trainig time on CPU as a function of Seq. Length (Sig. Input Size=6, Depth=3).}
    \label{fig:cpu_seqlen_train}
\end{figure}

\begin{figure}[ht]
    \centering
    \includegraphics[width=0.6\columnwidth]{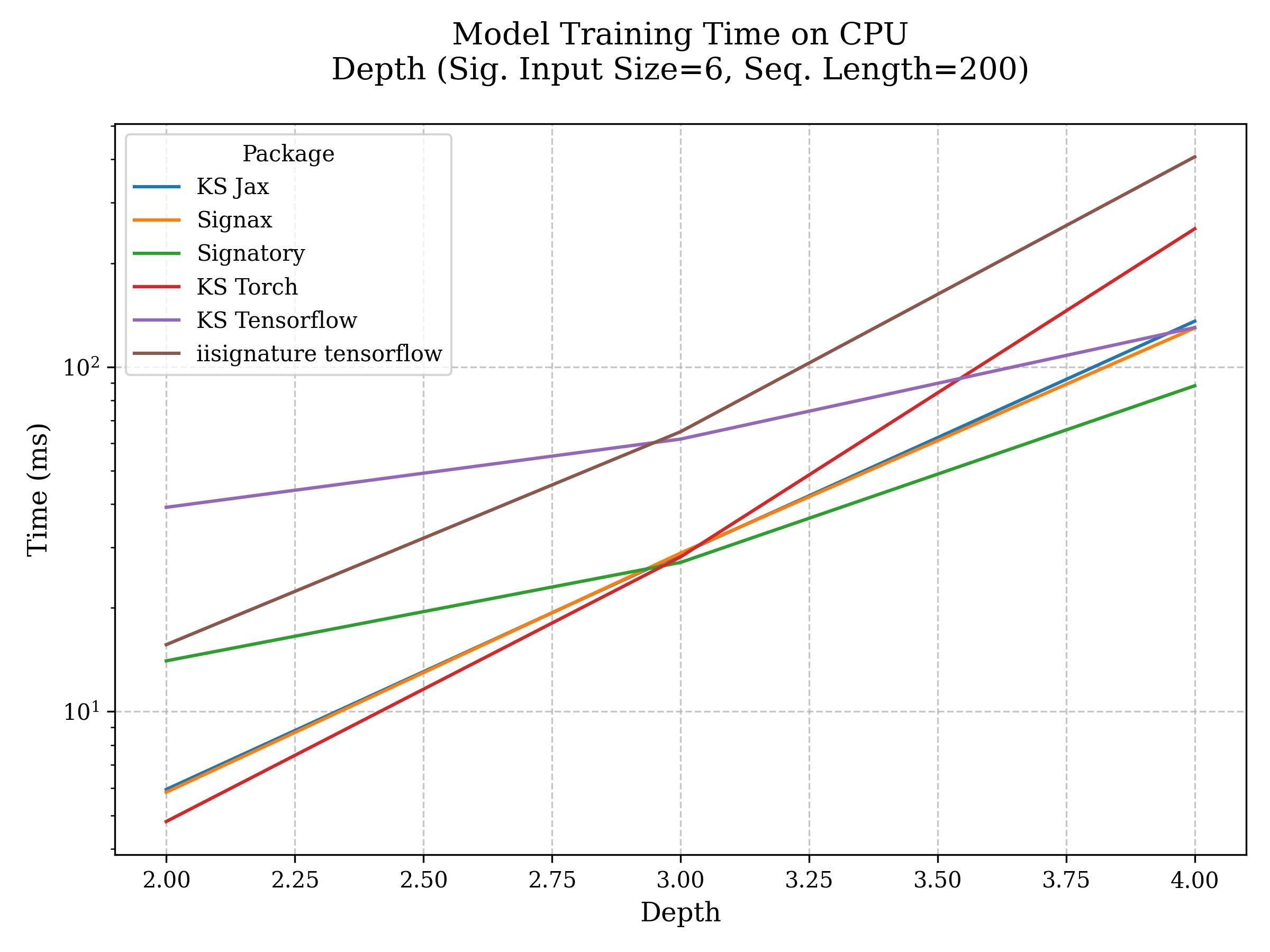}
    \caption{Model Trainig time on CPU as a function of Depth (Sig. Input Size=6, Seq. Length=200).}
    \label{fig:cpu_depth_train}
\end{figure}

Finally, we evaluated model training performance on CPU hardware. For CPU execution, Keras Sig automatically switches to a computation method similar to Signax's approach, with the primary difference being the use of Keras API rather than JAX primitives. The TensorFlow implementation in Keras Sig differs from other backends due to its handling of loop operations. While Keras provides a scan operation for implementing loops with controlled compiler optimization, its interaction with TensorFlow presents unique challenges. Loop unrolling—a compiler optimization technique that reduces loop overhead by replicating the loop body—becomes particularly problematic here. While scan operations typically allow fine-grained control over how much unrolling occurs, TensorFlow attempts to fully unroll loops by default. For long sequences, this aggressive unrolling strategy leads to prohibitively expensive compilation times, as the compiler attempts to create an expanded version of the entire loop. Consequently, while TensorFlow models can technically be compiled, the compilation overhead becomes impractical compared to training time, forcing us to use uncompiled models for TensorFlow benchmarks. For completeness, we included the iisignature TensorFlow 2 wrapper in our CPU benchmarks, running it within the Keras framework using the TensorFlow backend. Table \ref{tab:model_training_cpu} presents model training times across different configurations of signature input size, sequence length, and signature degree on CPU hardware.
\input{tables/model_training_cpu}
The results show that Keras Sig with JAX backend achieves performance nearly identical to Signax, which is expected given their shared computational foundation. A notable observation is that despite both JAX-based implementations and Signatory using similar algorithmic approaches, the JAX versions consistently outperform Signatory's optimized C++ implementation by margins ranging from 10\% to 300\%, depending on the configuration. This performance advantage demonstrates JAX's sophisticated optimization capabilities on CPU hardware. Performance across other backends exhibits interesting patterns. The PyTorch backend shows variable performance relative to JAX implementations, sometimes marginally outperforming and sometimes underperforming depending on specific parameter configurations. The TensorFlow backend presents a particularly interesting case: while it generally shows the slowest execution times among Keras Sig variants, for small signature sizes, the iisignature wrapper paradoxically demonstrates better performance than the native TensorFlow implementation. However, this advantage rapidly diminishes as signature complexity increases—with larger input sizes and higher degrees, the iisignature wrapper either becomes computationally prohibitive or fails to execute entirely. We provide the same visualization as before to highlight the effects. Figure \ref{fig:cpu_sig_size_train} illustrates training time as a function of signature input size, with sequence length fixed at 200 and signature degree set to 3. The results show the expected performance parity between Keras Sig with JAX backend and Signax. However, an interesting pattern emerges with larger signature input dimension: Signatory's performance begins to show advantages over the JAX-based implementations. This reversal of the GPU results highlights how JAX's XLA compilation optimizations, while extremely effective for GPU computation, may not fully match the efficiency of carefully optimized C++ code for CPU-specific operations when dealing with larger signature dimensions.
Figure \ref{fig:cpu_seqlen_train} examines how training time scales with sequence length, maintaining a fixed signature input size of 6 and signature degree of 3. The results reveal remarkably consistent scaling behavior across Signax, Keras Sig, and Signatory implementations. This uniformity in performance scaling is expected, as all implementations fundamentally employ similar computational strategies when processing sequences on CPU architecture, without the parallel processing advantages that distinguish GPU performance.
Figure \ref{fig:cpu_depth_train} analyzes the relationship between training time and signature degree, with signature input size fixed at 6 and sequence length at 200. The results mirror the patterns observed in our sequence length analysis, showing consistent scaling behavior across all implementations. This similarity in performance characteristics across different parameter variations reinforces our understanding that on CPU hardware, the fundamental computational approaches of all implementations converge to similar efficiency levels.
The model training experiments demonstrate several key findings about signature computation in deep learning environments. On GPU hardware, our optimized implementation shows substantial performance improvement, particularly for longer sequences, achieving up to 55\% reduction in training time compared to existing methods. However, these improvements are bounded by memory constraints when dealing with large signature dimensions, highlighting the trade-off between computational efficiency and memory usage. The backend comparison reveals important practical considerations: JAX-based implementations consistently deliver superior performance through effective XLA optimization, while PyTorch and TensorFlow implementations show varying degrees of efficiency. Notably, the inability to compile PyTorch models could become a significant limitation in more complex architectures. On CPU hardware, the performance landscape shifts considerably. The advantages of our GPU-optimized approach diminish, and all implementations demonstrate similar scaling characteristics. JAX-based implementations maintain a slight edge in most cases, though Signatory's optimized C++ code shows competitive performance, particularly with larger signature dimensions.

%% file: tables/signature_computation_gpu.tex
\begin{table}[!ht]
    \centering
    \caption{Signature Computation Time (ms) on GPU}
    \resizebox{\textwidth}{!}{%
        \begin{tabular}{rrrrrrrrr}
        \toprule
        Batch Size & Seq. length & Depth & KS Jax & KS GPU & KS Torch & KS Tensorflow & Signax & Signatory \\
        \midrule
        32 & 100 & 4 & 12.61 & 0.1312 & 1.11 & 16.49 & 0.6496 & 1.991 \\
        64 & 100 & 4 & 13.19 & 0.1424 & 1.109 & 16.29 & 0.6338 & 1.992 \\
        128 & 100 & 4 & 12.81 & 0.1785 & 1.159 & 16.39 & 0.656 & 2.015 \\
        256 & 100 & 4 & 15.0 & 0.1833 & 1.099 & 16.33 & 0.6906 & 2.696 \\
        512 & 100 & 4 & 14.0 & 0.2621 & 1.113 & 16.37 & 0.6789 & 3.211 \\
        128 & 50 & 4 & 13.58 & 0.1399 & 1.094 & 16.45 & 0.358 & 1.985 \\
        128 & 100 & 4 & 14.01 & 0.1437 & 1.09 & 16.38 & 0.6266 & 2.269 \\
        128 & 200 & 4 & 13.92 & 0.1798 & 1.09 & 16.31 & 1.194 & 3.565 \\
        128 & 500 & 4 & 14.31 & 0.5718 & 1.096 & 16.38 & 2.916 & 5.464 \\
        128 & 1000 & 4 & 15.79 & 0.5417 & 1.096 & 16.41 & 5.663 & 8.78 \\
        128 & 100 & 2 & 4.966 & 0.1689 & 0.44 & 5.917 & 0.659 & 0.9017 \\
        128 & 100 & 3 & 8.584 & 0.1712 & 0.722 & 10.5 & 0.8249 & 1.383 \\
        128 & 100 & 4 & 13.43 & 0.1485 & 1.086 & 16.39 & 0.6221 & 2.027 \\
        128 & 100 & 5 & 20.27 & 0.1585 & 1.516 & 23.69 & 0.7668 & 2.862 \\
        128 & 100 & 6 & 28.16 & 0.2903 & 2.024 & 32.91 & 1.229 & 3.919 \\
        \bottomrule
        \end{tabular}
        }
    \label{tab:sig_computation_gpu}
\end{table}

%% file: tables/signature_computation_cpu.tex
\begin{table}[!ht]
    \centering
    \caption{Signature Computation Time (ms) on CPU}
    \resizebox{\textwidth}{!}{%
        \begin{tabular}{rrrrrrrrrrr}
        \toprule
        Batch size & Seq. length & Depth & KS Jax & KS GPU & KS Torch & KS Tensorflow & Signax & Signatory & iisignature & esig \\
        \midrule
        32 & 100 & 4 & 143.5 & 2.032 & 1.472 & 214.4 & 0.2862 & 1.092 & 2.417 & 7.135 \\
        64 & 100 & 4 & 144.5 & 4.713 & 2.928 & 216.0 & 0.824 & 1.316 & 4.843 & 14.31 \\
        128 & 100 & 4 & 146.0 & 6.12 & 3.966 & 219.5 & 1.588 & 1.512 & 9.658 & 28.56 \\
        256 & 100 & 4 & 146.9 & 14.46 & 6.831 & 227.3 & 3.104 & 2.099 & 19.28 & 57.12 \\
        512 & 100 & 4 & 147.1 & 32.92 & 12.34 & 245.0 & 6.602 & 3.095 & 38.66 & 114.1 \\
        128 & 50 & 4 & 100.5 & 2.94 & 1.625 & 110.3 & 0.8289 & 1.001 & 4.774 & 19.32 \\
        128 & 100 & 4 & 151.8 & 7.009 & 2.113 & 226.1 & 2.035 & 1.697 & 9.653 & 28.63 \\
        128 & 200 & 4 & 43.53 & 14.1 & 4.548 & 453.2 & 2.262 & 3.092 & 19.44 & 47.27 \\
        128 & 500 & 4 & 45.99 & 45.64 & 14.61 & 1124.0 & 7.419 & 7.181 & 48.69 & 102.2 \\
        128 & 1000 & 4 & 132.6 & 75.98 & 32.26 & 2230.0 & 15.78 & 14.09 & 97.63 & 194.8 \\
        128 & 100 & 2 & 51.73 & 0.419 & 0.6012 & 73.93 & 0.1508 & 1.148 & 1.331 & 11.97 \\
        128 & 100 & 3 & 92.97 & 1.509 & 1.116 & 137.8 & 0.5496 & 1.276 & 3.805 & 15.74 \\
        128 & 100 & 4 & 156.9 & 6.745 & 1.879 & 223.1 & 1.773 & 1.528 & 9.649 & 28.73 \\
        128 & 100 & 5 & 217.8 & 34.51 & 5.25 & 347.7 & 4.18 & 2.149 & 25.81 & 75.72 \\
        128 & 100 & 6 & 321.6 & 96.95 & 25.09 & 520.6 & 10.46 & 3.952 & 75.45 & 268.5 \\
        \bottomrule
        \end{tabular}
    }
    \label{tab:sig_computation_cpu}
\end{table}

%% file: tables/model_training_gpu.tex
\begin{table}[!ht]
    \centering
    \caption{Model Training Time (seconds) on GPU}
    \resizebox{0.9\textwidth}{!}{%
        \begin{tabular}{rrrrrrrr}
        \toprule
        Seq. length & Sig input size & depth & KS Jax & Signax & Signatory & KS Torch & KS Tensorflow \\
        \midrule
        100 & 2 & 2 & 3.376 & 3.314 & 5.945 & 4.242 & 4.173 \\
        100 & 2 & 3 & 4.938 & 3.95 & 7.273 & 7.339 & 4.981 \\
        100 & 2 & 4 & 5.092 & 5.438 & 10.15 & 8.086 & 6.249 \\
        100 & 4 & 2 & 3.212 & 3.089 & 5.2 & 3.58 & 4.448 \\
        100 & 4 & 3 & 4.523 & 3.942 & 7.392 & 6.375 & 5.849 \\
        100 & 4 & 4 & 6.118 & 5.578 & 10.09 & 8.724 & 7.325 \\
        100 & 6 & 2 & 3.388 & 2.94 & 5.264 & 3.641 & 4.719 \\
        100 & 6 & 3 & 4.186 & 3.947 & 7.299 & 4.827 & 6.6 \\
        100 & 6 & 4 & 5.662 & 5.709 & 10.18 & 9.086 & 11.51 \\
        100 & 10 & 2 & 3.514 & 2.872 & 5.241 & 4.213 & 5.399 \\
        100 & 10 & 3 & 4.47 & 4.341 & 7.376 & 4.541 & 7.905 \\
        100 & 10 & 4 & 9.79 & 7.528 & 10.24 & 17.69 & 25.06 \\
        200 & 2 & 2 & 3.249 & 4.232 & 7.459 & 4.024 & 4.164 \\
        200 & 2 & 3 & 3.78 & 5.424 & 10.93 & 5.019 & 4.934 \\
        200 & 2 & 4 & 5.396 & 7.768 & 15.72 & 9.434 & 5.944 \\
        200 & 4 & 2 & 2.783 & 4.152 & 7.409 & 4.012 & 4.716 \\
        200 & 4 & 3 & 3.826 & 5.439 & 10.95 & 8.502 & 5.241 \\
        200 & 4 & 4 & 5.005 & 8.146 & 15.75 & 6.352 & 6.883 \\
        200 & 6 & 2 & 2.83 & 3.968 & 7.399 & 4.006 & 4.388 \\
        200 & 6 & 3 & 3.73 & 5.549 & 11.0 & 5.308 & 5.592 \\
        200 & 6 & 4 & 5.378 & 8.496 & 15.78 & 8.924 & 19.17 \\
        200 & 10 & 2 & 3.039 & 3.655 & 7.429 & 4.022 & 4.528 \\
        200 & 10 & 3 & 4.216 & 6.112 & 11.05 & 5.885 & 11.96 \\
        200 & 10 & 4 & 12.41 & 11.61 & 17.97 & 33.73 & 85.96 \\
        350 & 2 & 2 & 3.496 & 5.774 & 10.43 & 7.378 & 4.782 \\
        350 & 2 & 3 & 4.645 & 7.623 & 16.49 & 8.392 & 5.584 \\
        350 & 2 & 4 & 5.612 & 11.22 & 24.68 & 10.52 & 6.729 \\
        350 & 4 & 2 & 3.239 & 5.612 & 10.45 & 7.956 & 4.98 \\
        350 & 4 & 3 & 4.663 & 7.878 & 16.54 & 9.613 & 6.464 \\
        350 & 4 & 4 & 6.222 & 12.07 & 24.83 & 10.17 & 9.307 \\
        350 & 6 & 2 & 3.675 & 5.175 & 10.42 & 7.382 & 5.101 \\
        350 & 6 & 3 & 4.397 & 7.866 & 16.59 & 8.424 & 7.24 \\
        350 & 6 & 4 & 7.075 & 12.49 & 24.9 & 14.19 & 50.6 \\
        350 & 10 & 2 & 3.42 & 5.188 & 10.48 & 7.37 & 5.437 \\
        350 & 10 & 3 & 5.143 & 8.904 & 16.64 & 11.34 & 27.11 \\
        350 & 10 & 4 & 17.69 & 17.57 & 30.63 & 60.83 & 287.1 \\
        500 & 2 & 2 & 3.27 & 7.164 & 10.44 & 7.422 & 5.422 \\
        500 & 2 & 3 & 4.559 & 9.844 & 16.62 & 8.408 & 6.379 \\
        500 & 2 & 4 & 5.194 & 14.86 & 24.77 & 9.573 & 7.526 \\
        500 & 4 & 2 & 3.324 & 7.153 & 10.48 & 7.173 & 5.675 \\
        500 & 4 & 3 & 4.365 & 10.26 & 16.66 & 8.124 & 7.481 \\
        500 & 4 & 4 & 5.958 & 16.02 & 24.92 & 9.813 & 13.53 \\
        500 & 6 & 2 & 3.678 & 6.62 & 10.47 & 7.15 & 5.928 \\
        500 & 6 & 3 & 4.655 & 10.24 & 16.6 & 8.635 & 9.125 \\
        500 & 6 & 4 & 7.812 & 16.52 & 25.05 & 17.67 & 95.44 \\
        500 & 10 & 2 & 3.811 & 6.622 & 10.47 & 7.134 & 6.465 \\
        500 & 10 & 3 & 5.305 & 11.43 & 16.73 & 13.6 & 48.33 \\
        500 & 10 & 4 & 22.35 & 23.23 & 33.51 & 84.28 & 596.5 \\
        \bottomrule
        \end{tabular}
        
        }
    \label{tab:model_training_gpu}
\end{table}

%% file: tables/model_training_cpu.tex
\begin{table}[!ht]
    \centering
    \caption{Model Training Time (seconds) on CPU}
    \resizebox{0.9\textwidth}{!}{%
        \begin{tabular}{lllllllll}
        \toprule
        Seq. length & Sig input size & depth & KS Jax & Signax & Signatory & KS Torch & KS Tensorflow & iisignature Tensorflow \\
        \midrule
        100 & 2 & 2 & 2.834 & 1.641 & 6.856 & 3.753 & 13.08 & 5.697 \\
        100 & 2 & 3 & 5.316 & 3.481 & 8.727 & 5.36 & 24.18 & 8.139 \\
        100 & 2 & 4 & 11.56 & 5.793 & 10.96 & 7.621 & 38.97 & 12.23 \\
        100 & 4 & 2 & 2.511 & 2.184 & 7.923 & 3.749 & 14.01 & 7.419 \\
        100 & 4 & 3 & 6.791 & 4.301 & 10.6 & 5.69 & 23.98 & 17.0 \\
        100 & 4 & 4 & 25.19 & 13.93 & 19.52 & 18.64 & 41.96 & 52.95 \\
        100 & 6 & 2 & 4.168 & 3.609 & 7.964 & 3.847 & 14.76 & 9.35 \\
        100 & 6 & 3 & 15.77 & 15.52 & 14.74 & 9.537 & 26.07 & 35.6 \\
        100 & 6 & 4 & 65.97 & 62.51 & 45.43 & 120.0 & 58.31 & 206.5 \\
        100 & 10 & 2 & 8.734 & 8.457 & 9.697 & 4.272 & 18.46 & 15.93 \\
        100 & 10 & 3 & 47.65 & 50.22 & 30.4 & 83.58 & 37.05 & 126.1 \\
        100 & 10 & 4 & 351.7 & 235.1 & 208.4 & 890.9 & 273.0 & 1470.0 \\
        200 & 2 & 2 & 2.544 & 2.67 & 11.48 & 4.437 & 26.81 & 7.646 \\
        200 & 2 & 3 & 5.47 & 5.387 & 15.09 & 6.473 & 48.38 & 12.5 \\
        200 & 2 & 4 & 10.75 & 10.12 & 19.43 & 9.32 & 80.89 & 20.0 \\
        200 & 4 & 2 & 3.099 & 3.035 & 13.17 & 4.274 & 31.74 & 10.81 \\
        200 & 4 & 3 & 7.393 & 7.291 & 18.59 & 7.638 & 54.17 & 28.01 \\
        200 & 4 & 4 & 31.38 & 31.5 & 36.61 & 38.22 & 91.43 & 98.14 \\
        200 & 6 & 2 & 5.945 & 5.834 & 14.04 & 4.798 & 39.15 & 15.64 \\
        200 & 6 & 3 & 28.67 & 28.84 & 27.09 & 28.1 & 61.75 & 64.85 \\
        200 & 6 & 4 & 135.8 & 129.7 & 88.19 & 251.7 & 130.1 & 407.1 \\
        200 & 10 & 2 & 15.34 & 15.26 & 17.11 & 7.983 & 51.23 & 27.4 \\
        200 & 10 & 3 & 106.2 & 105.4 & 58.07 & 167.8 & 94.57 & 238.8 \\
        200 & 10 & 4 & 404.2 & 406.9 & 414.9 & 1748.0 & 596.3 & 2952.0 \\
        350 & 2 & 2 & 3.55 & 3.644 & 17.15 & 5.123 & 52.83 & 11.36 \\
        350 & 2 & 3 & 8.787 & 8.644 & 23.63 & 7.7 & 92.64 & 19.15 \\
        350 & 2 & 4 & 17.95 & 17.59 & 32.01 & 11.44 & 153.0 & 55.67 \\
        350 & 4 & 2 & 4.314 & 4.325 & 20.26 & 5.345 & 75.42 & 17.62 \\
        350 & 4 & 3 & 12.48 & 12.45 & 30.53 & 14.82 & 113.7 & 46.04 \\
        350 & 4 & 4 & 50.75 & 50.67 & 62.66 & 101.2 & 183.1 & 351.8 \\
        350 & 6 & 2 & 9.364 & 9.204 & 22.79 & 6.618 & 94.81 & 23.75 \\
        350 & 6 & 3 & 47.53 & 47.64 & 45.33 & 64.51 & 136.7 & 109.2 \\
        350 & 6 & 4 & 232.8 & 230.9 & 151.7 & 465.4 & 260.2 & 1523.0 \\
        350 & 10 & 2 & 25.19 & 25.18 & 27.65 & 16.28 & 133.2 & 44.59 \\
        350 & 10 & 3 & 176.3 & 175.9 & 100.0 & 302.9 & 206.6 & 416.0 \\
        350 & 10 & 4 & 656.8 & 653.6 & 723.7 & 3027.0 & 1079.0 & \_ \\
        500 & 2 & 2 & 4.623 & 4.678 & 24.53 & 5.825 & 91.13 & 26.26 \\
        500 & 2 & 3 & 11.76 & 11.77 & 33.38 & 9.148 & 145.8 & 45.63 \\
        500 & 2 & 4 & 25.17 & 25.47 & 43.45 & 14.34 & 232.2 & 80.74 \\
        500 & 4 & 2 & 5.708 & 5.757 & 29.94 & 6.684 & 132.9 & 38.08 \\
        500 & 4 & 3 & 17.28 & 17.33 & 42.85 & 28.64 & 191.4 & 122.2 \\
        500 & 4 & 4 & 63.76 & 62.61 & 89.04 & 153.5 & 287.1 & 505.0 \\
        500 & 6 & 2 & 12.63 & 12.75 & 30.25 & 9.513 & 180.2 & 56.89 \\
        500 & 6 & 3 & 65.35 & 64.81 & 64.34 & 100.5 & 232.2 & 318.2 \\
        500 & 6 & 4 & 327.9 & 327.5 & 215.2 & 674.5 & 409.1 & 2176.0 \\
        500 & 10 & 2 & 35.25 & 35.25 & 39.11 & 25.47 & 244.6 & 124.4 \\
        500 & 10 & 3 & 253.5 & 252.4 & 141.7 & 434.4 & 350.8 & 1285.0 \\
        500 & 10 & 4 & 905.0 & 904.9 & 1032.0 & 4354.0 & 1517.0 & \_ \\
        \bottomrule
        \end{tabular}
    }
    \label{tab:model_training_cpu}
\end{table}

%% file: 4_Conclusion.tex
\section{Conclusion}
This work introduces \textit{Keras Sig}, a pure Python implementation of the signature transform that addresses key challenges in integrating signature-based methods into modern deep learning workflows. Through our comprehensive evaluation, we have demonstrated that high-level tensor operations, when properly optimized for GPU architectures, can match or exceed the performance of traditional low-level implementations while offering greater flexibility and maintainability.Our GPU-optimized approach achieves significant performance improvements, showing up to 55\% reduction in training time compared to existing methods for long sequences, and 5 to 10-fold improvements in direct signature computation. These gains stem from our innovative reorganization of computations to maximize parallel processing through large matrix operations, although the benefits are bounded by memory constraints for large signature dimensions. The backend-agnostic design of \textit{Keras Sig}, enabled by the Keras 3 framework, offers several practical advantages. While JAX-based implementations consistently deliver the best performance through XLA optimization, our PyTorch and TensorFlow implementations remain competitive, providing users flexibility in their choice of deep learning framework. This flexibility, combined with pure Python implementation, significantly reduces the versioning and maintenance issues that have historically plagued signature computation libraries. However, our approach also reveals important trade-offs that practitioners should consider. The memory requirements of our GPU-optimized implementation can become significant for large signature dimensions or sequence lengths. Additionally, on CPU hardware, the performance advantages of our approach diminish, with all implementations showing similar scaling characteristics. These limitations highlight the importance of automatic backend selection in \textit{Keras Sig}, which ensures optimal performance across different hardware configurations. Looking forward, this work opens several promising directions for future research. The success of our GPU optimization strategy suggests potential for similar reorganizations of other sequential computations in deep learning. Additionally, the demonstrated benefits of high-level tensor operations over low-level C++ implementations may encourage a broader shift towards more maintainable, framework-agnostic implementations of mathematical tools in deep learning.Through its combination of performance, flexibility, and ease of use, \textit{Keras Sig} represents a significant step forward in making signature-based methods more accessible to the deep learning community, while establishing new standards for the implementation of mathematical tools in modern machine learning frameworks.